\documentclass[10pt,twocolumn,letterpaper]{article}

\usepackage[accsupp]{axessibility}
\usepackage{cvpr}              % To produce the CAMERA-READY version
\usepackage[dvipsnames]{xcolor}
\usepackage[utf8]{inputenc} % allow utf-8 input
\usepackage[T1]{fontenc}    % use 8-bit T1 fonts
\usepackage{url}            % simple URL typesetting
\usepackage{booktabs}       % professional-quality tables
\usepackage{amsfonts}       % blackboard math symbols
\usepackage{nicefrac}       % compact symbols for 1/2, etc.
\usepackage{microtype}      % microtypography
\usepackage{xcolor}         % colors
\usepackage{boldline}
\usepackage{arydshln}
\usepackage{multirow}
\usepackage{amsmath}
\usepackage{xspace}
\usepackage{caption}
\usepackage{longtable}
\usepackage{graphicx}
\usepackage{makecell}
\usepackage{pifont}
\usepackage{utfsym}

\makeatletter
\def\thickhline{%
	\noalign{\ifnum0=`}\fi\hrule \@height \thickarrayrulewidth \futurelet
	\reserved@a\@xthickhline}
\def\@xthickhline{\ifx\reserved@a\thickhline
	\vskip\doublerulesep
	\vskip-\thickarrayrulewidth
	\fi
	\ifnum0=`{\fi}}
\def\eg{\emph{e.g.}\xspace}

\makeatother

\definecolor{cvprblue}{rgb}{0.21,0.49,0.74}
\usepackage[pagebackref,breaklinks,colorlinks,citecolor=cvprblue]{hyperref}

\def\paperID{***} % *** Enter the Paper ID here
\def\confName{CVPR}
\def\confYear{2024}

\title{Amodal Ground Truth and Completion in the Wild}

\author{%
  Guanqi Zhan$^1$, Chuanxia Zheng$^1$, Weidi Xie$^{1,2}$, Andrew Zisserman$^1$\\
    $^1$VGG, University of Oxford, \hspace{5pt}
    $^2$CMIC, Shanghai Jiao Tong University\\
  \texttt{\{guanqi,cxzheng,weidi,az\}@robots.ox.ac.uk} \\
}
\begin{document}
\maketitle
\begin{abstract}

This paper studies amodal image segmentation: predicting entire object segmentation masks including both visible and invisible (occluded) parts.
In previous work, the amodal segmentation ground truth on real images is usually predicted by manual annotaton and thus is subjective. 
In contrast, we use 3D data to establish an automatic pipeline to determine authentic {\em ground truth} amodal masks for partially occluded objects in real images. This pipeline is used to construct an amodal completion {\em evaluation} benchmark, \emph{MP3D-Amodal}, consisting of a variety of object categories and labels. 
To better handle the amodal completion task in the wild, 
we explore two architecture variants: a two-stage model that first infers the occluder, followed by amodal mask completion; 
and a one-stage model that exploits the representation power of Stable Diffusion for amodal segmentation across many categories.
Without bells and whistles, our method achieves a new state-of-the-art performance on Amodal segmentation datasets that cover a large variety of objects, including COCOA and our new \emph{MP3D-Amodal} dataset. The dataset, model, and code are available at \url{https://www.robots.ox.ac.uk/~vgg/research/amodal/}.
\end{abstract}

\section{Introduction}

\begin{figure}[h]
	\centering
	\includegraphics[height=1.6 \linewidth]{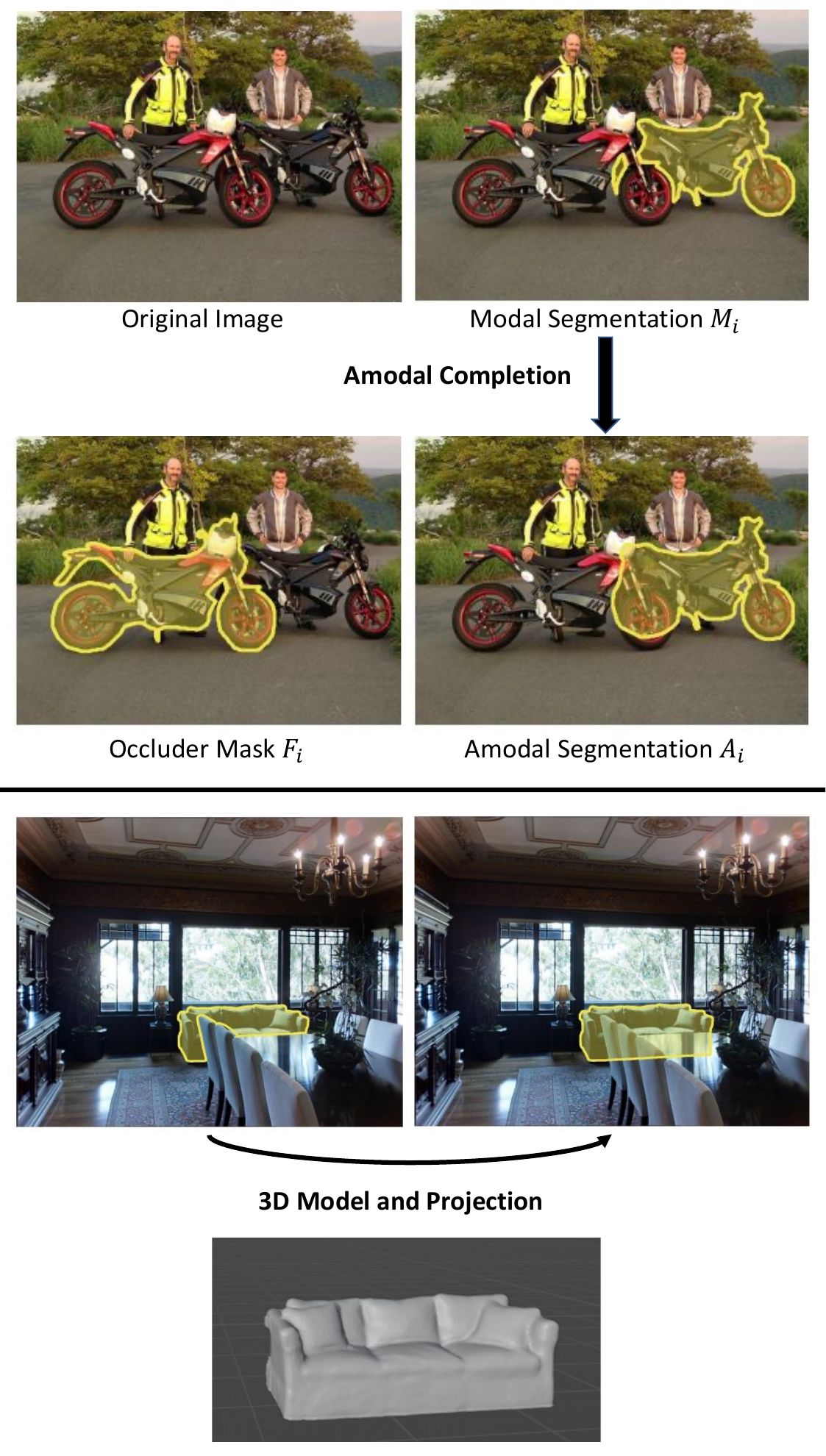}
	\vspace{-9pt}
\caption{\textbf{Amodal Ground Truth and Completion in the Wild.}
Top: The task of amodal completion is to predict the amodal mask $A_i$ for an object `$i$' in the image specified by the modal mask $M_i$ (here the object is the rear motorbike).
Previous methods~\cite{zhan2020self,nguyen2021weakly} require the mask of the occluder $F_i$ to be also provided to do the task;  but our goal is to predict the amodal mask when the occluder mask is {\em not} provided and the occluded object is of {\em any category}.
Bottom: We propose a novel method for generating amodal masks for real images: using 3D structure to
produce {\em ground truth} modal and amodal masks for object instances. The method is used to generate a ground truth evaluation dataset for real images.
}
\vspace{-30pt}
\label{fig:teaser}
\end{figure}

\begin{table*}[h]
    \centering
    \renewcommand{\arraystretch}{1.0}
    \tabcolsep=0.6 cm
    \footnotesize
    \begin{tabular}{@{}lccccc@{}}
    \toprule
        Dataset & Amodal GT & Image Type & \# Categories & \# Images & Type \\
    \midrule
         COCOA~\cite{zhu2017semantic}  & \ding{55} & Real &  Multiple  & 5,073 & General \\
         COCOA-cls~\cite{follmann2019learning}  & \ding{55} & Real &  80  & 3,499 & General \\ 
         KINS~\cite{qi2019amodal} & \ding{55} & Real & 8 & 14,991 & Vehicle \\
         DYCE~\cite{ehsani2018segan} & \checkmark & Synthetic & 79 & 5,500 & Indoor \\
         OLMD~\cite{Dhamo2019iccv} & \checkmark &  Synthetic & 40 & 13,000 & Indoor\\
         CSD~\cite{zheng2021visiting} & \checkmark &  Synthetic & 40 & 11,434 & Indoor \\
         D2SA~\cite{follmann2019learning} & \checkmark &  Synthetic & 60 & 5,600 & Industrial\\
         KITTI-360-APS~\cite{mohan2022amodal} & \ding{55} & Real & 17 & 61,168 & Vehicle \\
         BDD100K-APS~\cite{mohan2022amodal} & \ding{55} & Real & 16 & 202 & Vehicle \\
         WALT~\cite{reddy2022walt} & \checkmark & Real &  2 & 60,000 & Vehicle \\
         MUVA~\cite{li2023muva} & \checkmark & Synthetic &  20 & 26,406 & Shopping \\
    \midrule
        MP3D-Amodal (Ours) & \checkmark & Real & 427(40) & 10,883 & Indoor \\
    \bottomrule
    \end{tabular}
    \caption{\textbf{Comparison of Different Amodal Datasets}.
    Amodal GT: whether the dataset provides ground truth amodal annotations or is manually guessed. \# represents the number of the following types. Our MP3D-Amodal dataset (Sec.~\ref{sec:eval_data}) has 427 different semantic labels mapped to 40 different MatterPort categories. Note,  the WALT dataset consists of video frames from 12 camera viewpoints, mainly of vehicles moving. 
    }
    \vspace{-10pt}
    \label{table:compare_dataset}
\end{table*}

The vision community has rapidly improved instance segmentation
performance over the last few years with a succession of powerful models,
 such as Mask-RCNN~\cite{he2017mask},
Mask2Former~\cite{cheng2021mask2former}, and Seg-Anything
(SAM)~\cite{kirillov2023segment}.  However, despite this remarkable progress,  these models only provide pixel-level {\em modal} segmentations for objects in the images, \emph{i.e.}, the instance masks are for the \emph{visible} pixels. 
The models lack the human ability to infer the full extent of the object in an image, 
when it is partially occluded. The prediction of {\em amodal masks}, 
which covers the full extent of the object, 
can assist several downstream tasks including object detection~\cite{zhan2022tri},
smart image editing~\cite{zhan2020self,ling2020variational,xu2023amodal}, 3D reconstruction from a single image~\cite{kar2015category,zou2020wacv,cmrKanazawa18,wu2023magicpony,ozguroglu2024pix2gestalt}, object permanence in video segmentation~\cite{xie2022segmenting,van2023tracking,hsieh2023tracking}, predicting support relationships between objects~\cite{silberman2012indoor,Zhuo2017IndoorSP},
and more generally for planning in manipulation and navigational tasks where reasoning on the full extent of the object could lead to improvements~\cite{kellman2013perceptual,amodal_robot_manipulation,kim2016planning,inagaki2019detecting,wu2023learning}. 

Predicting amodal masks for objects in 2D images is challenging: 
this is because real scenes contain a vast collection of
different types of objects, resulting in complex occlusions when they are projected to 2D images from a 3D physical world.  
To accurately complete the amodal shape of partially occluded objects requires a prediction of occlusion relations (in order to infer if and where the object is partially occluded), as well as predicting the shape of the occluded regions.
This challenge is also reflected in the type of amodal datasets available which are mostly synthetic -- for real images, amodal masks are almost always `imagined' by human annotators providing the contour of the amodal mask, and there is no dataset available to evaluate amodal completions with \emph{authentic} ground truth annotations for a large variety of object categories (see Table~\ref{table:compare_dataset}).

In this paper, our first contribution is to provide a new amodal benchmark evaluation dataset based on authentic ground truth amodal segmentation for real images, 
and covering a large variety of objects.
The new dataset is called \emph{MP3D-Amodal}, and the amodal mask is obtained by projecting the 3D structure of  objects in the scene onto the image (Figure~\ref{fig:teaser} bottom). We build the dataset from MatterPort3D~\cite{Matterport3D} that has both 3D structure and real images of indoor scenes.
The dataset and generation method is described in Sec.~\ref{sec:eval_data}.

In most prior work, amodal completion algorithms required the occluder mask to be specified~\cite{zhan2020self,nguyen2021weakly} (Figure~\ref{fig:teaser} top).
Our second contribution is to propose two architecture variants that do not require the occluder mask to be supplied:  {\em OccAmodal}, a two-stage model that first infers the occluder, followed by amodal mask compeletion; and {\em SDAmodal}, a one-stage model that uses the features of a pre-trained Stable Diffusion network, benefiting from its strong outpainting capabilities. 
These architectures are described in Sec.~\ref{sec:architecture}.

We achieve state-of-the-art amodal completion performance on both the public COCOA~\cite{zhu2017semantic} dataset, and on our own \emph{MP3D-Amodal} benchmark. In particular, the one-stage model, {\em SDAmodal}, benefiting from the pre-trained Stable Diffusion model, is able to generalize to another dataset with objects from a different domain and different categories, demonstrating class-agnostic completion. Taken together, the handling of situations where occluder masks are not provided and the class-agnostic domain generalization, moves amodal completions towards an `in the wild' capability.

\begin{figure*}[t]
	\centering
	\includegraphics[height=0.40\linewidth]{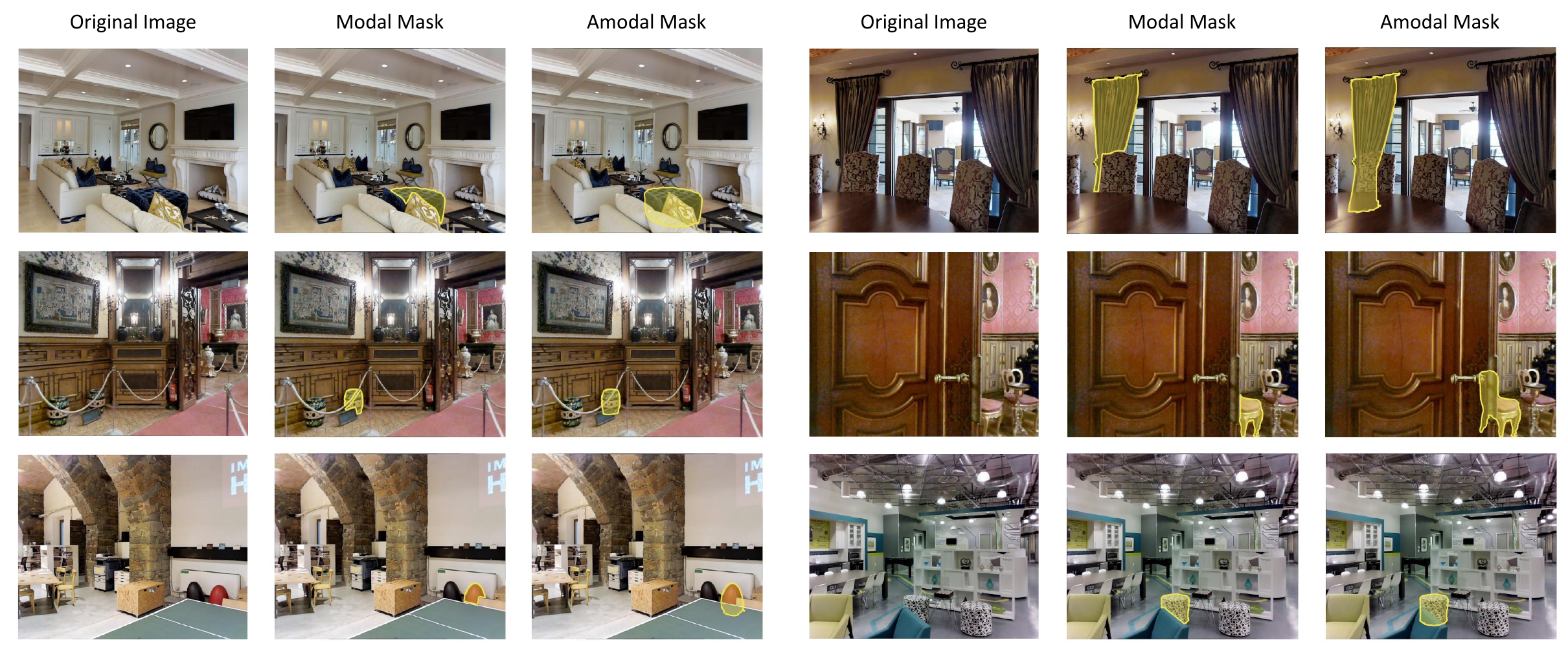}
\caption{\textbf{Samples from the \emph{MP3D-Amodal} Dataset.} For each sample, the original image 
is shown together with the generated modal and amodal masks. 
}
\label{fig:mp3d_example}
\vspace{-10pt}
\end{figure*}

\section{Related Work}

\noindent {\bf Amodal Datasets.} 
In the literature, there have been continuous efforts on creating datasets for amodal segmentation, for example, COCOA~\cite{zhu2017semantic, follmann2019learning} builds on COCO~\cite{lin2014coco}, KINS~\cite{qi2019amodal} builds on KITTI~\cite{geiger2012we}. However, the ground truth amodal masks for both of these datasets are created based on the 2D images, thus inevitably requiring human imagination for the occluded regions. To improve the quality of ground truth amodal mask, the DYCE~\cite{ehsani2018segan}, OLMD~\cite{Dhamo2019iccv}, CSD~\cite{zheng2021visiting} and MUVA~\cite{li2023muva} datasets were created by rendering the whole scene and corresponding individual intact objects using synthetic 3D models. 
WALT~\cite{reddy2022walt} collected objective amodal masks via time-lapse imagery, 
but their objects are limited to cars and humans.
Table~\ref{table:compare_dataset} provides a summary of the datasets currently available.
In contrast, we are the first to collect a complex dataset that provides authentic amodal ground truth for the occluded objects of a large variety of categories in real scenes.

\vspace{2pt} \noindent {\bf Amodal Instance Segmentation.} 
Classical instance segmentation methods~\cite{o2015learning,he2017mask,chen2019hybrid,cai2019cascade,cheng2021mask2former,kirillov2023segment} mainly focus on segmenting \emph{visible} pixels, 
while amodal instance segmentation~\cite{li2016amodal} aims to detect the objects as a whole, \emph{i.e.}, both \emph{visible} and \emph{invisible} parts. 
These methods are usually trained on images~\cite{qi2019amodal,follmann2019learning,zheng2021visiting,xiao2021amodal,ke2021deep,mohan2022amodal,sun2022amodal,tran2022aisformer,li20222d,gao2023coarse,li2023gin} with manually annotated ground truth amodal masks in a fully-supervised manner. 
However, these methods are trained on datasets with limited number of object classes, \emph{e.g.}, 80 categories for COCOA-cls, and are class dependent. 

\vspace{2pt} \noindent {\bf Amodal Completion} is conceptually similar to amodal instance segmentation, except that here the \emph{visible} mask for the target object is already provided as input.
Most existing methods~\cite{zhan2020self,nguyen2021weakly} assume the occluder mask is provided and
cannot handle the situation where the object is occluded by an unknown occluder, 
\emph{i.e.}, the occluder mask is not provided or the occluder is difficult to define. The methods are trained on COCOA, which covers a large variety of categories, and are more class-agnostic than methods trained on COCOA-cls.
Another work~\cite{ling2020variational} uses VAE~\cite{kingma2013auto} to model the task of amodal completion, but can only handle limited categories in driving scenes.

\begin{figure*}[h]
	\centering
	\includegraphics[height=0.39 \linewidth]{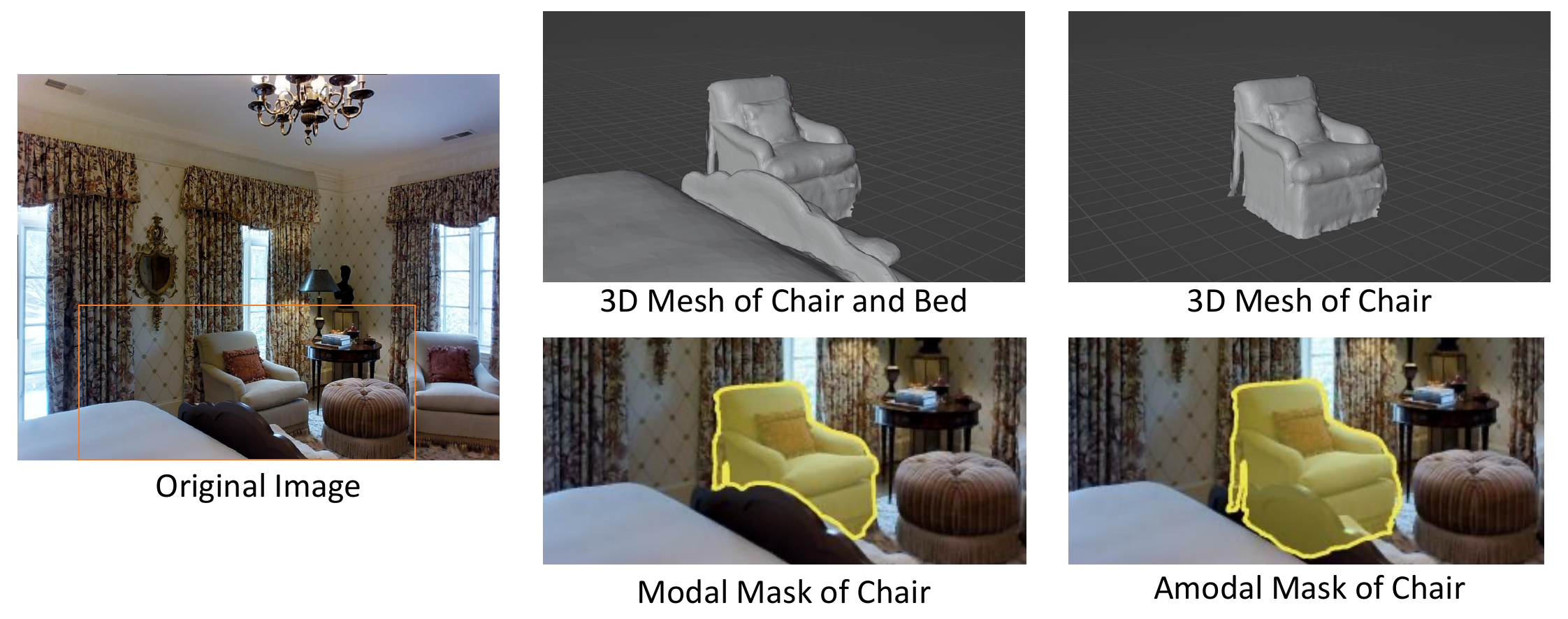}
\caption{\textbf{Automated Generation of the \emph{MP3D-Amodal} Ground Truth Dataset.} 
The dataset is automatically generated from the
MatterPort3D~\cite{Matterport3D} dataset, and provides ground truth
modal and amodal masks for objects in real
images. The generation process is illustrated here for the chair and proceeds in two steps: first, 
modal and amodal masks in a particular image are obtained for each object by projecting 
the object's 3D mesh individually (for the amodal mask), and also by projecting the 3D mesh of all objects (for
the modal mask). In this example, the 3D mesh of the bed occludes the chair when projected into the image.
In the second step, an object is selected for the dataset if 
 its amodal mask is larger than its modal
mask by a threshold. 
In this case the chair is selected, but other objects such as the stool would not be selected since
it is not occluded by other objects in this viewpoint, and so their modal and amodal masks would be the same.
}
\label{fig:eval_dataset}
% \vspace{-5pt}
\end{figure*}

\begin{figure}[h]
		\centering
		\includegraphics[trim=0.5cm 0cm 0.5cm 0cm, 
       height= 1.1 \linewidth    
        ]{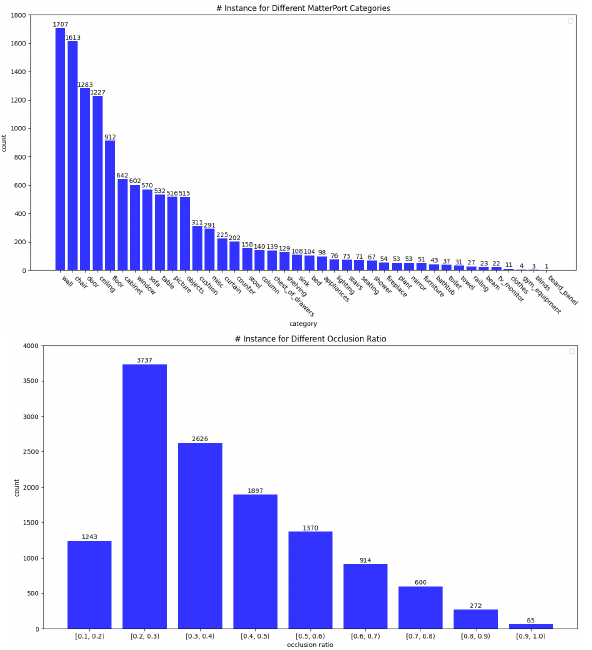}
		% \vspace*{-2mm}
		\caption{
  \textbf{Distributions of the \emph{MP3D-Amodal} Dataset} in terms of the number of instances for each MatterPort category, and the number of instances for different occlusion ratios. 
    }
		\label{fig:sup_mp3d_data_hist}
  % \vspace{-10pt}
\end{figure}

\section{The MP3D-Amodal Dataset}
\label{sec:eval_data}

In this section we describe the new amodal dataset \emph{MP3D-Amodal}, that is constructed from the MatterPort3D~\cite{Matterport3D} dataset. 
We first overview the contents of the dataset in Sec.~\ref{sec:overview_mp3d_dataset}, and then describe our method of generating ground truth amodal masks on real images from 3D data in Sec.~\ref{sec:generate_mp3d_dataset}.

\begin{table}[h]
\setlength{\tabcolsep}{2.5pt}
\footnotesize
\centering
\begin{tabular}{l|c|c|c|c|c}
\toprule 

Split & \# Scenes & \# Images & \# Instances & \makecell{\# MatterPort \\ Categories} & \makecell{\# Semantic \\ Labels} \\
\midrule
Training & 4 & 1,100 & 1,283 & 35 & 130 \\
Evaluation & 86 & 9,783 & 11,441 & 40 & 385 \\
Total & 90 & 10,883 & 12,724 & 40 & 427 \\

\bottomrule
\end{tabular}
\caption{\textbf{Statistics of the generated \emph{MP3D-Amodal} dataset.}
Each instance has a semantic label as annotated in the MatterPort3D dataset, which is also mapped to a more general MatterPort category.
Across the training and evaluation splits, there are 88 semantic labels in common, and 297 semantic labels in the evaluation split but not in the training split. 
} 
\vspace{-10pt}
\label{table:eval_data_statistics}
\end{table}

\subsection{An Overview of the Dataset}
\label{sec:overview_mp3d_dataset}

The dataset contains 12,724 annotated amodal ground truth masks for over 10,883 real images. Since it is built from the MatterPort dataset, 
we use the classifications inherited from that dataset, 
where objects are described by their {\em category} and {\em semantic labels}.
Note that, categories are more coarse-grained than semantic labels and one category may contain several different semantic labels, \emph{e.g.}, the category `chair' contains semantic labels `dining chair', `sofa chair' and `armchair'; and the category `appliances' contains `refrigerator', `oven', and `washing machine'.

Table~\ref{table:eval_data_statistics} gives the details for dataset splits. 
To have a better and more comprehensive evaluation, we make the evaluation split to have  more scenes than the training split.
Across the training and evaluation splits, there are 88 semantic labels in common, and 297 semantic labels in the evaluation split but not in the training split. 
A small part of the collected dataset is reserved for training, as this allows some domain adaptation for an algorithm. The scenes of the training set are disjoint from those of the evaluation set.

Samples from the dataset are displayed in Figure~\ref{fig:mp3d_example}. 
The dataset contains diverse range of objects, with some categories not in the `general' COCOA dataset, \emph{e.g.,} the example in the bottom left of Figure~\ref{fig:mp3d_example} is a novel category. More examples of the dataset are in the appendix.

Figure~\ref{fig:sup_mp3d_data_hist} visualizes the distributions of the dataset in terms of the number of
instances for each MatterPort category, and the number of instances for different occlusion ratios, where the occlusion ratio is the proportion of the the object that is occluded (the difference between amodal and modal masks, divided by the area of the amodal mask). It is evident that there is a wide range of occlusion ratios, from slightly occluded to severely occluded.

\begin{figure*}[t]
	\centering
	\includegraphics[height=0.27 \linewidth]{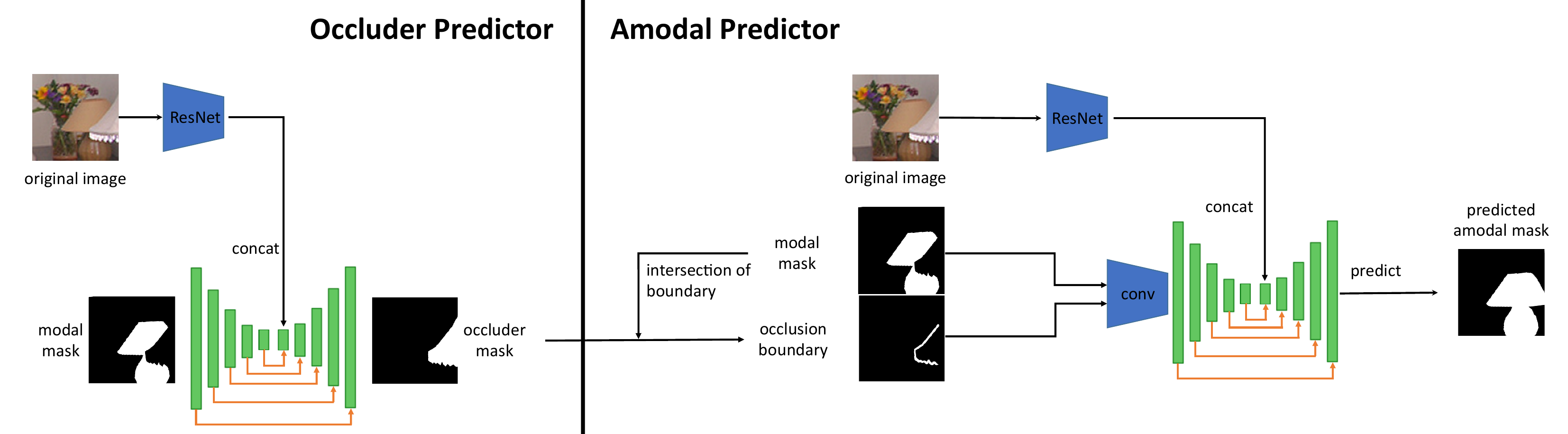}
 \vspace{-5pt}
\caption{
\textbf{Two-Stage Architecture (OccAmodal) for Amodal Prediction.} {\em Left}: A lightweight U-Net based architecture is used to predict the occluder mask for each object. 
{\em Right}: The amodal predictor  takes the predicted occluder mask, together with the modal mask and image as input to predict the amodal segmentation mask. 
}
\label{fig:architecture}
\vspace{-10pt}
\end{figure*}

% \vspace{-8pt}
\subsection{Generating Amodal Ground Truth from 3D}
\label{sec:generate_mp3d_dataset}
% \vspace{5pt}

We exploit the MatterPort3D~\cite{Matterport3D} dataset, 
that is equipped with two essential elements: 
a 3D mesh for each object instance in the scene, 
and real images (and associated cameras) of the scene. 
In the following we detail the procedure for automatically generating amodal and modal masks of individual objects. The process is illustrated in Figure~\ref{fig:eval_dataset}.

\vspace{3pt}\noindent {\bf Modal Mask Generation.} 
For a particular scene, 
we obtain 2D instance segmentations (a modal mask
for each object)  by projecting all objects with their
instance labels together onto the image with the associated camera. 
If $M_{i}$ and $O_{i}$ denote the modal mask and 3D mesh of the $i$-th object, 
$\Phi$ refers to the projection of 3D meshes to the camera space, then the modal masks of the image are:
\begin{align}
\label{eq:generate_modal}
    \{M_{1}, M_{2}, \dots, M_{n}\} = \Phi(O_{1} \cup O_{2} \cup \dots \cup O_{n})
\end{align}

\vspace{3pt}\noindent {\bf Amodal Mask Generation.} 
The amodal mask $A_{i}$ for each object $i$ is simply obtained by projecting each object to the camera separately:
\begin{align}
\label{eq:generate_amodal}
    \{A_{i}\} = \Phi(O_{i}), ~~~~~~\forall i \in \{1,2,3,\dots,n\}
\end{align}

\noindent {\bf Occluded Object Selection.} 
Then partially occluded objects are identified and selected by comparing the modal and
amodal mask.  If the amodal mask of the object is larger than the
modal mask, then there must be something occluding the object, and that
object's modal and amodal masks are candidates to be included in the dataset.
Here we automatically include objects with $S(A_{i}) > 1.2 ~S(M_{i})$, \emph{i.e.}, the area of its amodal mask is more than 1.2 times larger than its modal mask. Take the chair in
Figure~\ref{fig:eval_dataset} as an example, we first generate its
modal and amodal mask using Equations~\ref{eq:generate_modal} 
and~\ref{eq:generate_amodal}. Because the amodal mask of the chair is
larger than its modal mask (it is occluded by the bed), we
select the chair in the dataset.
In this way, we have an automatic method to collect ground truth amodal masks for occluded objects in real images without any manual guessing.

\vspace{3pt}\noindent {\bf Manual Selection.} 
However, not all generated modal and amodal masks are of very good
quality as the 3D meshes in MatterPort3D can be incomplete or noisy
sometimes. We thus apply a manual selection stage,
where human annotators inspect and select the pairs with good-quality
modal and amodal masks. Bad quality examples due to problems of MatterPort3D are categorized (\emph{e.g.,} the modal mask does not contain all visible parts of the object, or the amodal mask is noisy / incomplete) and shown to the human annotators. The manual selection is described in full detail in the appendix.

\section{Architectures for Amodal Prediction}
\label{sec:architecture}

\begin{figure}[t]
	\centering
	\includegraphics[height=0.46 \linewidth]{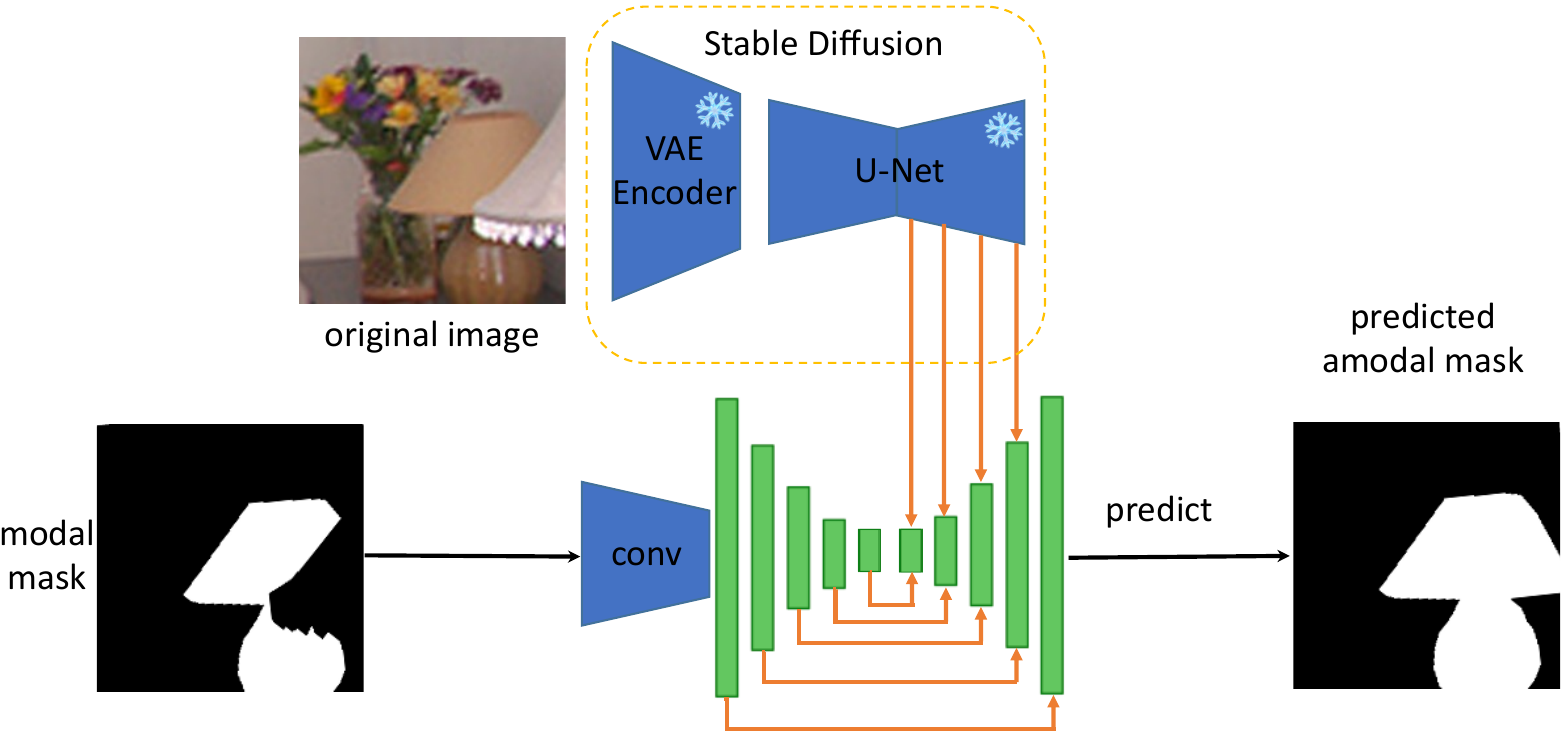}
 \vspace{-5pt}
\caption{
\textbf{One-Stage Architecture (SDAmodal) for Amodal Prediction.} The image is fed into a pre-trained Stable Diffusion model to get multi-scale representations containing occlusion information. The image and modal mask features are concatenated and forwarded to multiple decoding layers for amodal prediction. The Stable Diffusion model is frozen during  training.
}
\label{fig:architecture_sd}
\vspace{-10pt}
\end{figure}

Given a single image $\mathcal{I} \in \mathbb{R}^{H \times W \times 3}$ and its corresponding modal (\emph{visible}) mask $M_i \in \mathbb{R}^{H \times W}$ for the $i$-th object, our goal is to predict the amodal (\emph{full}) mask for the object, $A_i \in \mathbb{R}^{H \times W}$. Specifically, we explore two architecture variants:
\begin{itemize}
\item 
A two-stage architecture, as shown in Figure~\ref{fig:architecture},  
consisting of an {\bf occluder predictor} to first estimate the occluder mask, 
followed by an {\bf amodal predictor} to infer the amodal mask,
given the modal mask, estimated occluder, and image.
\item 
A one-stage architecture, as shown in Figure~\ref{fig:architecture_sd}, 
that exploits the strong representation power of the pre-trained stable diffusion model, 
and adapts it to infer the amodal mask from the given image and modal mask.
\end{itemize}

\subsection{Two-Stage Architecture -- OccAmodal}

\noindent {\bf Occluder Predictor.}
Occlusion in an image occurs when an object hides a part of another object, referred to as occluder and occludee respectively. 
For amodal completion, having the occluder's mask can largely simplify the task,
as it provides information on which parts of one specified object should be completed~\cite{zhan2020self,nguyen2021weakly}.
In existing works~\cite{zhan2020self,nguyen2021weakly}, 
the occluder mask is often considered as a prior, and is directly fed into the model as input. One obvious limitation, however, is that the occluder mask can be unavailable at inference time. For example, in large-scale datasets, \eg COCO~\cite{lin2014coco} or LVIS~\cite{gupta2019lvis}, not all objects in an image are annotated, resulting in a failure of amodal completion in existing works~\cite{zhan2020self,nguyen2021weakly}, \emph{i.e.,} they cannot expand the modal mask at all if the segmentation of occluder mask is not annotated and provided.
Here, instead of relying on an a-prior occluder mask, 
we consider a two-stage architecture, that first infers the occluder mask from the given image and the target object's modal mask, and then generates an amodal completion with the occluder mask as guidance. Specifically, as shown in Figure~\ref{fig:architecture}~(left), 
the occluder predictor takes the original image and the object's modal mask as input, 
to the ResNet and U-Net encoder respectively, and is then concatenated and upsampled to generate the prediction of the occluder mask for the object, $F_i = \Psi_{\{\textsc{OCP}\}}(\mathcal{I}, M_i)$,
where $F_i \in \mathbb{R}^{H \times W}$ denotes the binary mask of occluder,
that can be completely empty~(no occluder), or with the union of all occluders.

\vspace{3pt}
\noindent {\bf Amodal Predictor.}
Given the mask of the predicted  occluder, we compute the occlusion boundary~($B_i$), 
 between the modal mask and occluder mask.
We then feed the input image, object's modal mask, and occlusion boundary to an amodal predictor, as shown in Figure~\ref{fig:architecture}~(right),
similar to existing work~\cite{nguyen2021weakly}.
In detail, both the input modal mask~($M_i$) and occlusion boundary~($B_i$) are concatenated,
and input to a U-Net for encoding and decoding, with skip connections.
Additionally, we also encode the input image with a ResNet, and inject it into the U-Net's bottleneck layer, providing visual conditioning for amodal completion. 
We denote the procedure as : $A_i = \Psi_{\{\textsc{AMP}\}}(\mathcal{I}, M_i, F_i)$

\subsection{One-stage Architecture -- SDAmodal}

In recent literature, generative models based on diffusion have demonstrated the ability to generate photorealistic images, with seemingly correct geometry, object semantics and shapes. Here, we  investigate the possibility of exploiting the visual features in diffusion models for amodal completion -- after all, the task is mainly about understanding the shape of certain objects. Specifically, as shown in Figure~\ref{fig:architecture_sd}, 
we feed the image into a pre-trained Stable Diffusion model, 
and add noise onto the latent features after the VAE encoder.
We extract the multi-scale features from the decoding layers of the U-Net at time step 181
following the investigation about occlusion in~\cite{Zhan2023physd}. 
Then we concatenate the Stable Diffusion features with multi-scale features of the modal mask, 
and forward them to multiple decoding layers to generate the amodal mask prediction,
the procedure can be denoted as : $A_i = \Psi_{\{\textsc{SD}\}}(\mathcal{I}, M_i)$.

\subsection{Training}

Training the first stage of OccAmodal requires ground truth occluder masks, while training both the second stage of OccAmodal and the SDAmodal requires the ground truth amodal masks. Both COCOA and our MP3D-Amodal provide ground truth amodal masks while only COCOA provides ground truth occluder masks.

\vspace{3pt}\noindent {\bf OccAmodal.}
For training of the occluder predictor, the occlusion relationships annotated in COCOA~\cite{zhu2017semantic} are used to obtain the ground truth occluder mask, and then the pixel-level prediction of occluder mask is trained via cross-entropy loss. For training of the amodal predictor, the amodal mask prediction is supervised by the ground truth amodal mask (provided by COCOA or MP3D-Amodal) via an Uncertainty Weighted Segmentation Loss as mentioned in~\cite{nguyen2021weakly}.

\vspace{3pt}\noindent {\bf SDAmodal.}
For training of the Stable Diffusion based architecture, the amodal mask prediction is supervised by the ground truth amodal mask (provided by COCOA or MP3D-Amodal) via a cross-entropy loss.

\section{Experiments}
\label{sec:exp}

\begin{table*}
    \noindent\begin{minipage}[htbp]{0.48\textwidth}%

  \centering
  \footnotesize
  \setlength\tabcolsep{4pt}
  \begin{tabular}{ccccc}
\toprule 
\multirow{2}{*}{ID} & \multirow{2}{*}{Occluder Predictor} & \multirow{2}{*}{\makecell{Final Skip Connection}} & \multicolumn{2}{c}{COCOA} \\
\cline{4-5}
& & &  mIOU & mIOU-inv \\
\midrule
$\mathbb{A}$ &  &  & 69.9 & 0.006   \\ 
$\mathbb{B}$ & \checkmark &  & 88.4 & 64.4   \\ 
$\mathbb{C}$ & \checkmark & \checkmark & \textbf{89.4} & \textbf{66.2}   \\ 
\bottomrule
\end{tabular}
\captionof{table}{\textbf{Ablation Study of OccAmodal.} Setting  $\mathbb{A}$ is the setting of ASBU~\cite{nguyen2021weakly}. All models are trained on COCOA.}
\label{table:ablation_1}
\end{minipage}
\hspace{0.4cm}
\begin{minipage}[htbp]{0.48\textwidth}%
  \hspace{0.2cm}
  \footnotesize
  \setlength\tabcolsep{4pt}
  
  \begin{tabular}{cccccc}
\toprule 
\multirow{2}{*}{ID} & \multirow{2}{*}{SD Backbone} & \multirow{2}{*}{\makecell{Multi-Scale \\ SD Feature}} & \multirow{2}{*}{\makecell{Final Skip \\ Connection}} & \multicolumn{2}{c}{COCOA} \\
\cline{5-6}
& & & & mIOU & mIOU-inv \\
\midrule
$\mathbb{A}$ &  &  &  & 88.0 & 63.8   \\ 
$\mathbb{B}$ & \checkmark &  &  & 89.4 & 69.2   \\ 
$\mathbb{C}$ & \checkmark & \checkmark &  & 89.6 & 69.8   \\ 
$\mathbb{D}$ &  &  & \checkmark & 89.2 & 66.4   \\ 
$\mathbb{E}$ & \checkmark &  & \checkmark & 90.5 & 71.1   \\ 
$\mathbb{F}$ & \checkmark & \checkmark & \checkmark & \textbf{90.7} & \textbf{71.6}   \\ 
\bottomrule
\end{tabular}
\captionof{table}{\textbf{Ablation Study of SDAmodal.} Setting  $\mathbb{A}$ is Deocclusion~(Single Stage) in ~\cite{zhan2020self}. All models are trained on COCOA.}
\label{table:ablation_2}
\end{minipage}
\end{table*}

\begin{table*}[t]
\setlength{\tabcolsep}{6pt}
\footnotesize
\centering

\begin{tabular}{cccccccc}
\toprule 
\multirow{2}{*}{ID} & \multirow{2}{*}{Comments} & \multirow{2}{*}{Occluder Mask Provided} & \multicolumn{2}{c}{COCOA} && \multicolumn{2}{c}{MP3D-Amodal} \\
\cline{4-5}\cline{7-8}
& & & mIOU & mIOU-inv && mIOU & mIOU-inv \\
\midrule
$\mathbb{A}$ & Deocclusion(Two Stage)~\cite{zhan2020self} & \checkmark  & 88.2 & 65.3 && - & -   \\ 
$\mathbb{B}$ & ASBU~\cite{nguyen2021weakly}(reproduced) & \checkmark  & 88.9 & 65.3 && - & -   \\ 
$\mathbb{C}$ & ASBU~\cite{nguyen2021weakly}(reported) & \checkmark  & 89.9 & - && - & -   \\ 
\midrule
$\mathbb{D}$ & Deocclusion~(Two Stage)~\cite{zhan2020self} &  & 69.9 & 0.006 && 64.4 & 0.004   \\ 
$\mathbb{E}$ & ASBU~\cite{nguyen2021weakly} &   & 69.9 & 0.006 && 64.4 & 0.004   \\ 
$\mathbb{F}$ & Deocclusion~(Single Stage)~\cite{zhan2020self} &  & 88.0 & 63.8 && 72.4 & 28.0   \\ 
$\mathbb{G}$ & OccAmodal &   & 89.4 & 66.2 && 72.9 & 27.5   \\ 
$\mathbb{H}$ & SDAmodal &   & \textbf{90.7} & \textbf{71.6} && \textbf{76.4} & \textbf{38.5}   \\ 
\bottomrule
\end{tabular}
\vspace{-5pt}
\caption{\textbf{Comparison with State-of-the-Art Amodal Completion Methods}. Our SDAmodal model achieves the new state-of-the-art performance for amodal completion over a larger variety of categories. All models are trained on COCOA, and evaluated on both COCOA and MP3D-Amodal. 
}
\label{table:compare_sota_1}
\vspace{-10pt}
\end{table*}

\vspace{8pt}
\subsection{Experimental Details}
\label{sec:dataset_impl}

\noindent {\bf Datasets and Implementation Details.} 
We employ both COCOA~\cite{zhu2017semantic} and our collected MP3D-Amodal~(Section~\ref{sec:eval_data}) for training and evaluating our models. 
To ensure a fair comparison, we use the same training setting as in~\cite{zhan2020self,nguyen2021weakly}, which employs SGD with momentum, 
sets the learning rate to be $1e^{-3}$, and trains the model for 56K iterations with a batch size of 32. 
Models are trained on A6000 / A40 GPUs. More training details are given in the appendix. 

\vspace{2pt}\noindent {\bf Baselines.} 
We compare with two existing amodal completion models~\cite{zhan2020self,nguyen2021weakly}, 
of which~\cite{nguyen2021weakly} is the latest state-of-the-art method for amodal completion.
\cite{zhan2020self} has both one-stage and two-stage architectures, which we denote as Deocclusion~(Single Stage) and Deocclusion~(Two Stage). The default architecture in~\cite{zhan2020self} is Deocclusion~(Two Stage), while Deocclusion~(Single Stage) uses a ResNet to encode the input image and concatenate it with the features in the U-Net decoder (similar to Figure~\ref{fig:architecture} left). 
Additionally, in the appendix,  we  compare with 
recent amodal instance segmentation methods such as VRSP~\cite{xiao2021amodal}, A3D~\cite{li20222d}, AISformer~\cite{tran2022aisformer}, C2F-Seg~\cite{gao2023coarse} and GIN~\cite{li2023gin}.

\vspace{2pt}\noindent {\bf Evaluation.} 
Following~\cite{zhan2020self,nguyen2021weakly},
we compute mIOU between the ground truth and predicted amodal mask. 
Additionally, mIOU-inv is also used, which refers to the mIOU for only the occluded regions.

\begin{table*}[ht]
\setlength{\tabcolsep}{6pt}
\footnotesize
\centering

\begin{tabular}{ccccccccc}
\toprule 
\multirow{2}{*}{ID} & \multirow{2}{*}{Architecture} & \multirow{2}{*}{COCOA} & \multirow{2}{*}{MP3D-Amodal} & \multicolumn{2}{c}{COCOA} && \multicolumn{2}{c}{MP3D-Amodal} \\
\cline{5-6}\cline{8-9}
& & & & mIOU & mIOU-inv && mIOU & mIOU-inv \\
\midrule
$\mathbb{A}$ & OccAmodal & \checkmark  &  & \textbf{89.4} & 66.2 && 72.9 & 27.5   \\ 
$\mathbb{B}$ & OccAmodal & \checkmark & \checkmark & \textbf{89.4} & \textbf{66.4} && \textbf{73.8} & \textbf{29.6}  \\ 
\midrule
$\mathbb{C}$ & SDAmodal & \checkmark  &  & \textbf{90.7} & \textbf{71.6} && 76.4 & 38.5   \\ 
$\mathbb{D}$ & SDAmodal & \checkmark & \checkmark  & \textbf{90.7} & \textbf{71.6}  && \textbf{81.8} & \textbf{53.7}  \\ 
\bottomrule
\end{tabular}
% \vspace{-5pt}
\caption{\textbf{Effectiveness of Different Training Data}. 
The performance of both models are boosted on MP3D-Amodal if extra training data from MP3D-Amodal is used.
}
\vspace{-10pt}
\label{table:different_train_data}
\end{table*}

\subsection{Ablation Study of Different Architectures}

In Table~\ref{table:ablation_1}, we ablate the importance of the occluder predictor and the number of skip connections for the OccAmodal architecture.
As can be seen, the occluder mask is crucial for amodal mask prediction. This is eveident from the results of Setting $\mathbb{A}$, achieving only 69.9 mIOU on COCOA. 
In comparison, when the predicted occluder mask is incorporated, amodal completion can be boosted to 88.4 mIOU~(Setting $\mathbb{B}$) on COCOA, and the performance is further boosted when we include a skip connection for the final layer of the U-Net (Setting $\mathbb{C}$). In the architecture of~\cite{nguyen2021weakly} there are only 4 skip connections and we are adding the fifth.
In Table~\ref{table:ablation_2}, we ablate variations on the SDAmodal architecture. Replacing the original ResNet image encoder with the Stable Diffusion backbone brings a significant boost (+1.4/+5.4 in terms of mIOU and mIOU-inv for Setting $\mathbb{A}$ to $\mathbb{B}$, +1.3/+4.7 in terms of mIOU and mIOU-inv for Setting $\mathbb{D}$ to $\mathbb{E}$). 
If multiple layers of Stable Diffusion features at different resolutions are fed into the model (as shown in Figure~\ref{fig:architecture_sd}) the performance is higher than if only a single layer feature is used (the second layer of the Stable Diffusion U-Net as in~\cite{Zhan2023physd})~(comparing Setting $\mathbb{B}$/$\mathbb{C}$ and $\mathbb{E}$/$\mathbb{F}$).
The performance can also be improved by adding a final layer skip connection for the U-Net (comparing Setting $\mathbb{A}$/$\mathbb{D}$,  $\mathbb{B}$/$\mathbb{E}$ and $\mathbb{C}$/$\mathbb{F}$).
According to~\cite{Zhan2023physd}, the features of other pre-trained models such as DINO~\cite{caron2021dino,oquab2023dinov2} and CLIP~\cite{Radford2021clip,software_openclip} perform worse than Stable Diffusion features on ``occlusion'' task. 
We have further trained our model using DINO and CLIP features. The results are given in the appendix, validating the superiority of Stable Diffusion features.

\begin{figure*}%[h]
	\centering
	\includegraphics[height=0.55 \linewidth]{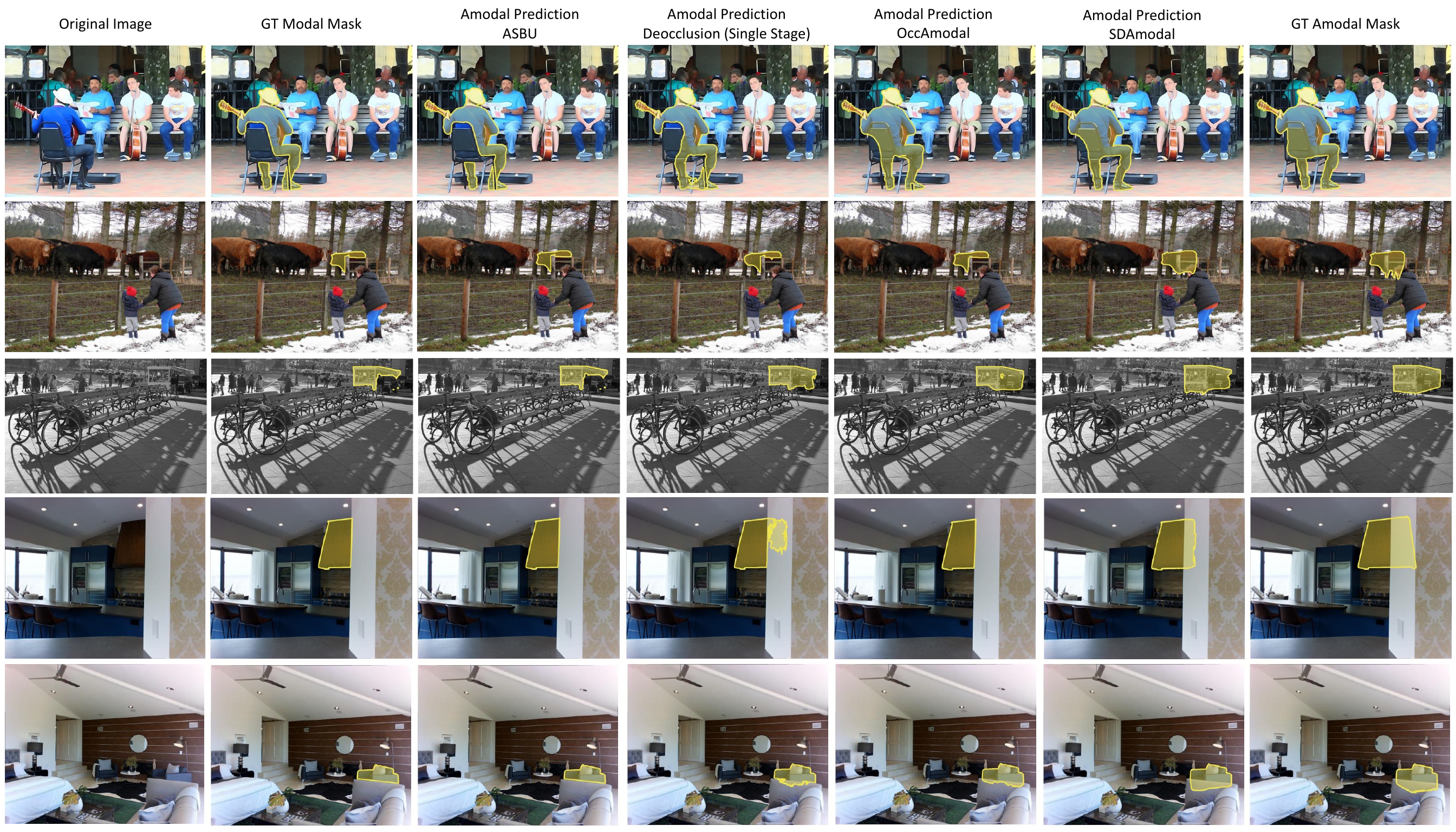}
 \vspace{-5pt}
\caption{\textbf{Qualitative Comparison on COCOA and MP3D-Amodal}. 
COCOA: Rows 1, 2 and 3; MP3D-Amodal: Rows 4 and 5. Please see the text for more discussion. More qualitative examples are provided in the appendix.
}
\label{fig:qualitative}
\vspace{-10pt}
\end{figure*}

\subsection{Comparison with State-of-the-Art}
\label{sec:compare_sota}

We compare our method with previous amodal completion state-of-the-art methods, Deocclusion~\cite{zhan2020self} and ASBU~\cite{nguyen2021weakly},
on both COCOA and MP3D-Amodal.
Note that, ASBU~\cite{nguyen2021weakly} and Deocclusion~(Two Stage)~\cite{zhan2020self} require the occluder masks provided, 
while in both of the architectures we propose, the occluder masks are not necessary.
The comparisons are given in Table~\ref{table:compare_sota_1}. We can make the following observations: 
(1) SDAmodal outperforms the previous state-of-the-art methods (Setting $\mathbb{C}$ and $\mathbb{H}$) even if the occluder mask is not provided for SDAmodal, but is for previous methods; 
(2) When the occluder mask is not provided, previous 
methods Deocclusion~(Two Stage)~\cite{zhan2020self} and ASBU~\cite{nguyen2021weakly} cannot expand the modal mask of the object and  achieve poor performance for amodal completion (Setting $\mathbb{D}$ and $\mathbb{E}$). In comparison, OccAmodal (Setting $\mathbb{G}$), where the occluder mask is generated by our occluder predictor, has a high performance, demonstrating the effectiveness of the occluder mask prediction module. 
(3) Even though SDAmodal is only trained on COCOA, the Stable Diffusion backbone  efficiently boosts the performance not only on COCOA, but also {\em zero-shot generalized} to MP3D-Amodal which contains objects from different domains and categories (compare Settings $\mathbb{F}$ and $\mathbb{H}$ where the difference is +4.0/+10.5 in terms of mIOU and mIOU-inv).
Comparison with recent Amodal Instance Segmentation methods are given in the appendix.

\subsection{Effectiveness of Different Training Data}
\label{sec:ablation}

Table~\ref{table:different_train_data} shows the effectiveness of training with extra data from our MP3D-Amodal training split. 
Both OccAmodal and SDAmodal improve performance on MP3D-Amodal when they are also trained with MP3D-Amodal and there is no performance deterioration on COCOA.

\subsection{Qualitative Results}
\label{sec:qualitative}

In Figure~\ref{fig:qualitative}, we show a qualitative comparison of different amodal completion methods on both the COCOA and MP3D-Amodal datasets. 
We observe that ASBU~\cite{nguyen2021weakly} faces limitations in expanding the modal mask when the occluder mask is not provided (Column 3). Deocclusion~(Single Stage) can partially complete the amodal mask when the occluder mask is not available but the prediction quality is not good (Column 4).
In contrast, our models, especially SDAmodal, can handle the situation where the occluder mask is not provided and significantly improve the accuracy of amodal mask predictions (Columns 5 and 6), even when the object to complete is from a different domain (Rows 4 and 5) when only trained on COCOA. 
More qualitative results are in the appendix.

\section{Conclusion and Extensions}
\label{sec:conclusion}
By utilising real 3D data, we have proposed an automatic pipeline to generate ground truth amodal masks for
occluded objects in real images, and used this to create a new
amodal segmentation evaluation benckmark for a large variety of
instances. The pipeline has been applied to the MatterPort3D dataset, but can be applied to other
suitable datasets such as ScanNet~\cite{dai2017scannet} that have real images together with the 3D structure for objects in the scene.
Furthermore,  we have developed two new state-of-the-art methods for
amodal completion \emph{in the wild} --  \emph{i.e.}, capable of handling situations where the occluder is unknown or undefined, and for a wide variety of object classes. The models, with a lightweight occluder predictor and Stable Diffusion representations,
achieve superior performance on different domains and object categories.

\vspace{-10pt}
\paragraph{Acknowledgements. } 
This research is supported by EPSRC Programme Grant VisualAI EP$\slash$T028572$\slash$1, a Royal Society
Research Professorship RP$\backslash$R1$\backslash$191132, an AWS credit funding, a China Oxford Scholarship and ERC-CoG UNION 101001212.
We thank Max Bain, Emmanuelle Bourigault, Abhishek Dutta, Kai Han, Tengda Han, Joao Henriques, Jaesung Huh, Tomas Jakab, Dominik Kloepfer, David Miguel Susano Pinto, Prasanna Sridhar, Ashish Thandavan, Vadim Tschernezki, and Yan Xia for their particular help; and Yash Bhalgat, Minghao Chen, Subhabrata Choudhury, Shu Ishida, Prajwal Kondajji Renukananda, Ragav Sachdeva, Sagar Vaze, and Chuhan Zhang from the Visual Geometry Group for general discussions on the project. 
We also thank Rajan from Elancer and his team, for their huge assistance with annotation.

{
    \small
    \bibliographystyle{ieeenat_fullname}
    \bibliography{main}
}

\appendix

\clearpage
\section*{Appendix}
\section{Additional Implementation Details}
\label{sec:sup_exp}

In this section, we provide additional details about our experiments: Section~\ref{sec:sup_train_detail} for more details about training our model, and Section~\ref{sec:sup_miou_inv} for a demonstration of mIOU-inv.

\subsection{Training Details}
\label{sec:sup_train_detail}

In Table~\ref{table:different_train_data} of the main paper, 
where different experiments use different extra training data, the extra training data (\emph{i.e.}, MP3D-Amodal) is usually mixed with the default training data (\emph{i.e.}, COCOA) in the random sampling process at a possibility ratio of 10\% : 90\%.

While training our two-stage models, including OccAmodal, 
in addition to using the amodal ground truth in COCOA, 
we also apply the data augmentation (artificial occlusion) as mentioned in ~\cite{zhan2020self,nguyen2021weakly} to enable a better performance of the trained models.

\subsection{mIOU vs.\ mIOU-inv}
\label{sec:sup_miou_inv}

\begin{figure}[h]
		\centering
		\includegraphics[trim=0.5cm 0cm 0.5cm 0cm, height=0.5 \linewidth]{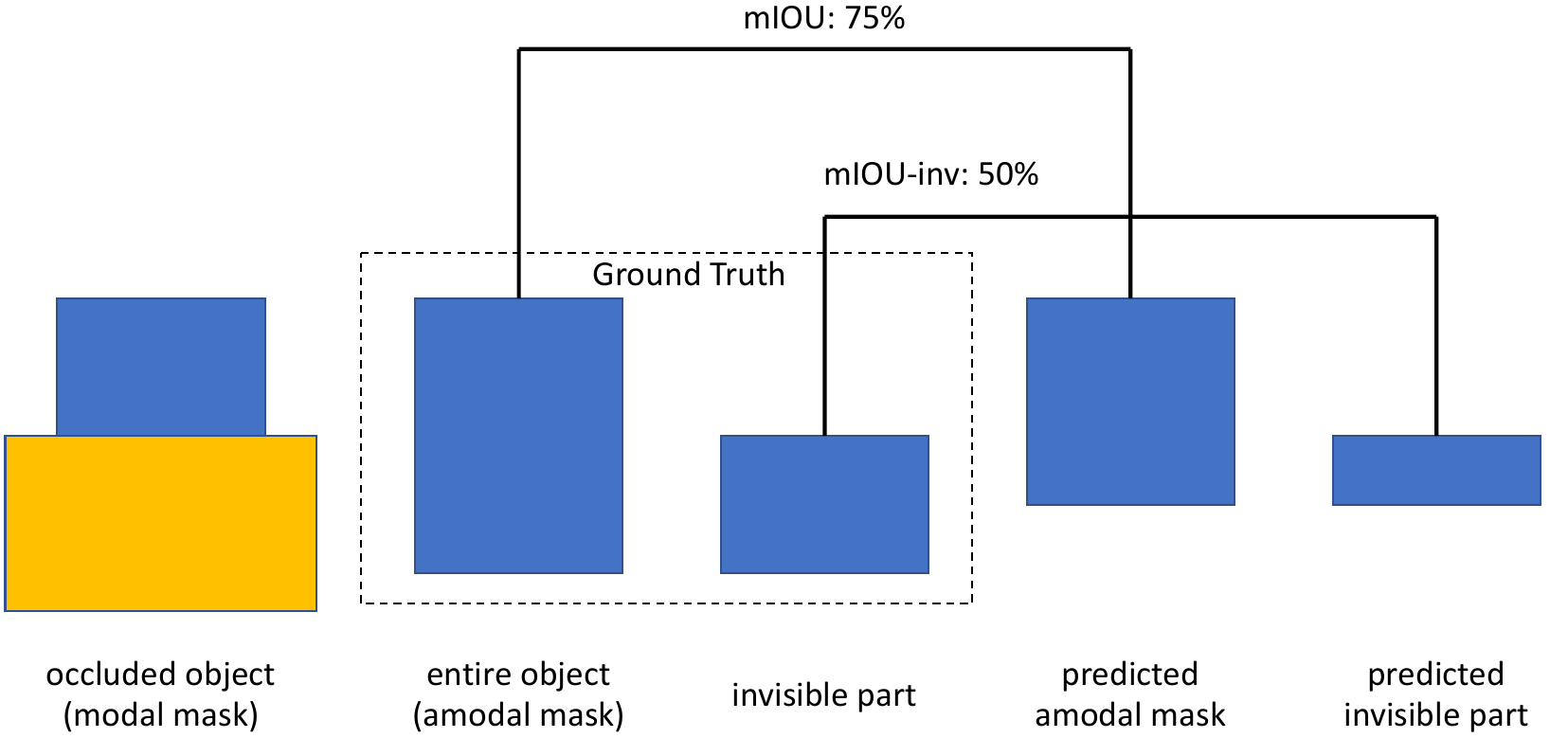}
		\caption{\textbf{mIOU vs.\ mIOU-inv}. 
  mIOU-inv is the mIOU of predicted invisible(occluded) part and ground truth invisible part, which is a more sensitive evaluation metric to evaluate the performance of amodal completion. For this example, the ground truth modal mask and amodal mask of the blue rectangle are shown in the first and second columns, and we can see 50\% of the blue rectangle is occluded by the yellow rectangle. The predicted amodal mask extends the original modal mask from 50\% to 75\%, and the mIOU is 75\% here. But when calculated the mIOU-inv, it is only 50\% because the amodal predictor only extends 50\% of the occluded region. Even if the predicted amodal mask does not extend the modal mask at all in this example, the mIOU is still 50\% while the mIOU-inv should be 0\%.}
		\label{fig:sup_miou}
\end{figure}

In Section~\ref{sec:exp} of the main paper, 
we adopt two evaluation metrics, mIOU and mIOU-inv. 
mIOU is the average of the IOU between the predicted amodal mask and ground truth amodal mask, while mIOU-inv is the average of the IOU between the predicted and ground truth amodal mask outside the original modal mask area. 
As Figure~\ref{fig:sup_miou} shows, mIOU-inv reflects the quality of predicted amodal mask in the original occluded area, which is a more sensitive evaluation metric than mIOU.

\section{Comparison with Amodal Instance Segmentation Methods}

As mentioned in Section~\ref{sec:exp} of the main paper, we also compare our models with recent amodal instance segmentation methods. 
We test our models on the COCOA-cls~\cite{follmann2019learning} benchmark, which is a subset of COCOA containing objects from only 80 COCO categories, and compare the mIOU and mIOU-inv metrics with the numbers reported in the recent amodal instance segmentation papers (state-of-the-art amodal instance segmentation methods like~\cite{gao2023coarse} have not made the code and model publicly available). 

In Table~\ref{table:compare_sota_2}, we can observe that both OccAmodal and SDAmodal
achieve superior performance than other amodal instance segmentation methods in terms of mIOU and mIOU-inv on COCOA-cls. 

\begin{table}[h]
\setlength{\tabcolsep}{6pt}
\footnotesize
\centering
\begin{tabular}{cccc}
\toprule 
\multirow{2}{*}{ID} & \multirow{2}{*}{Comments} & \multicolumn{2}{c}{COCOA-cls}  \\
\cline{3-4}
& & mIOU & mIOU-inv\\
\midrule

$\mathbb{A}$ & A3D~\cite{li20222d} & 67.4 & -   \\ 
$\mathbb{B}$ & GIN~\cite{li2023gin} & 72.4 & -   \\ 
$\mathbb{C}$ & AISformer~\cite{tran2022aisformer} & 72.7 & 13.8   \\ 
$\mathbb{D}$ & VRSP~\cite{xiao2021amodal} & 79.0 & 22.9   \\ 
$\mathbb{E}$ & C2F-Seg~\cite{gao2023coarse} & 87.1 & 36.6   \\ 
$\mathbb{F}$ & OccAmodal & 92.0 & 47.8   \\ 
$\mathbb{G}$ & SDAmodal & \textbf{93.5} & \textbf{59.6}   \\ 
\bottomrule
\end{tabular}
\caption{\textbf{Compare with State-of-the-Art Amodal Instance Segmentation Methods on COCOA-cls~\cite{follmann2019learning}}. Our methods achieve superior performance in terms of mIOU and mIOU-inv on COCOA-cls.
}
\label{table:compare_sota_2}
\end{table}

\section{Comparing Stable Diffusion Features with Other Large Pre-Trained Models}

As mentioned in Section~\ref{sec:architecture} of the main paper, in Table~\ref{table:dif_feat} we use the features from different large pre-trained models such as OpenCLIP~\cite{Radford2021clip,software_openclip}, DINOv1~\cite{caron2021dino} and DINOv2~\cite{oquab2023dinov2} in our one-stage SDAmodal architecture. It can be observed Stable Diffusion features perform the best among all features.

\begin{table}[h]
\setlength{\tabcolsep}{6pt}
\footnotesize
\centering
\begin{tabular}{lccccc}
\toprule 
\multirow{2}{*}{Feature} & \multicolumn{2}{c}{COCOA} && \multicolumn{2}{c}{MP3D-Amodal} \\
\cline{2-3}\cline{5-6}
 & mIOU & mIOU-inv && mIOU & mIOU-inv \\
\midrule
OpenCLIP & 88.3 & 65.2 && 73.3 & 31.0   \\ 
DINOv1 & 89.0 & 67.4 && 74.5 & 34.3   \\ 
DINOv2 & 90.3 & 70.4 && 76.3 & 38.1   \\ 
Stable Diffusion & \textbf{90.7} & \textbf{71.6} && \textbf{76.4} & \textbf{38.5}   \\ 
\bottomrule
\end{tabular}
\caption{\textbf{Amodal completion performance using different features in the One-stage architecture. }}
\label{table:dif_feat}
\end{table}

\section{More Qualitative Amodal Prediction Examples}
\label{sec:sup_qualitative}

\begin{figure*}[!htb]
        
		\centering
		\includegraphics[trim=0.5cm 0cm 0.5cm 0cm, height=1.17 \linewidth]{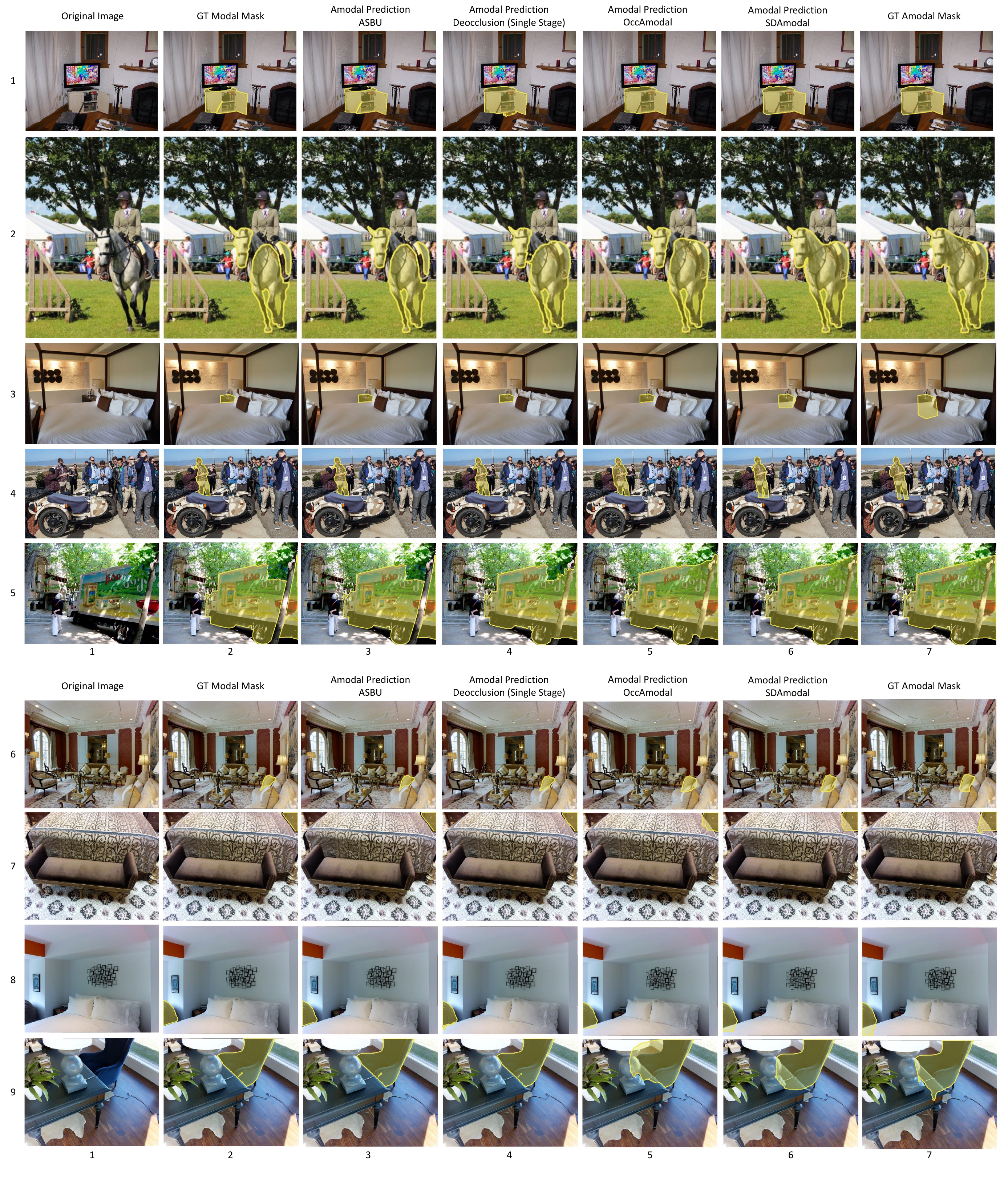}
		\vspace{-5mm}
		\caption{\textbf{More Qualitative Results Comparing Different Amodal Completion Methods}. Rows 1-5: COCOA; Rows 6-9: MP3D-Amodal. 
  It can be observed that ASBU~\cite{nguyen2021weakly} faces limitations in expanding the modal mask when the occluder mask is not provided (Column 3). Deocclusion~(Single Stage) can partially complete the amodal mask when the occluder mask is not available but the prediction quality is not good (Column 4).
In contrast, our models, especially SDAmodal, can handle the situation where the occluder mask is not provided and significantly improve the accuracy of amodal mask predictions (Columns 5 and 6), even when the object to complete is from a different domain (Rows 6-9) when only trained on COCOA.
  }
		\label{fig:sup_qualitative_pred_occluder}
 		\vspace{-4mm}
\end{figure*}

Figure~\ref{fig:sup_qualitative_pred_occluder} displays more qualitative examples comparing our models and previous amodal completion methods.

\clearpage

\section{The Amodal Evaluation Ground Truth Dataset MP3D-Amodal}
\label{sec:sup_eval_data}

More statistics of \emph{MP3D-Amodal} are given in Section~\ref{sec:sup_mp3d_stat},  and more examples of the \emph{MP3D-Amodal} are shown in Section~\ref{sec:sup_mp3d_example}.

\subsection{MP3D-Amodal Dataset Statistics}
\label{sec:sup_mp3d_stat}

\begin{figure*}[!htb]
		\centering
		\includegraphics[trim=0.5cm 0cm 0.5cm 0cm, height=0.18\linewidth]{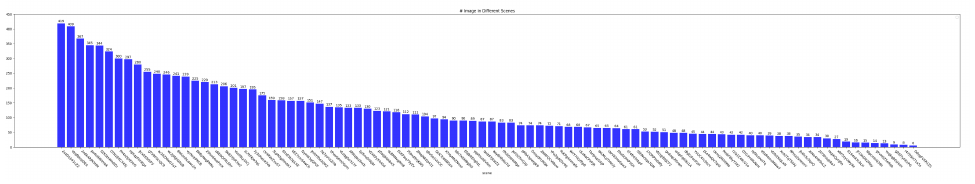}
		\caption{\textbf{Statistics of our collected MP3D-Amodal Dataset} in terms of the number of images in each MP3D scene. Please zoom in for the details. 
  }
		\label{fig:sup_mp3d_data_hist_2}
\end{figure*}

In Section~\ref{sec:overview_mp3d_dataset} of the main paper we show statistics of our collected MP3D-Amodal dataset in terms of number of instances for each MatterPort category and number of instances for different occlusion rates.
Additionally, Figure~\ref{fig:sup_mp3d_data_hist_2} shows more statistics of our collected MP3D-Amodal dataset, \emph{i.e.}, the number of images in each MP3D scene. It can be observed that our collected instances come from 90 different MatterPort3D scenes. For some scenes the quality of the 3D mesh is better so that we could collect more instances, while for scenes whose 3D mesh is not of good quality, we only collect instances with good quality masks so there are fewer samples. The difference in number also results from different number of original images for different MatterPort3D scenes.

\subsection{More Examples of the MP3D-Amodal Dataset}
\label{sec:sup_mp3d_example}

\begin{figure*}[!htb]
		\centering
		\includegraphics[trim=0.5cm 0cm 0.5cm 0cm, height=1.2 \linewidth]{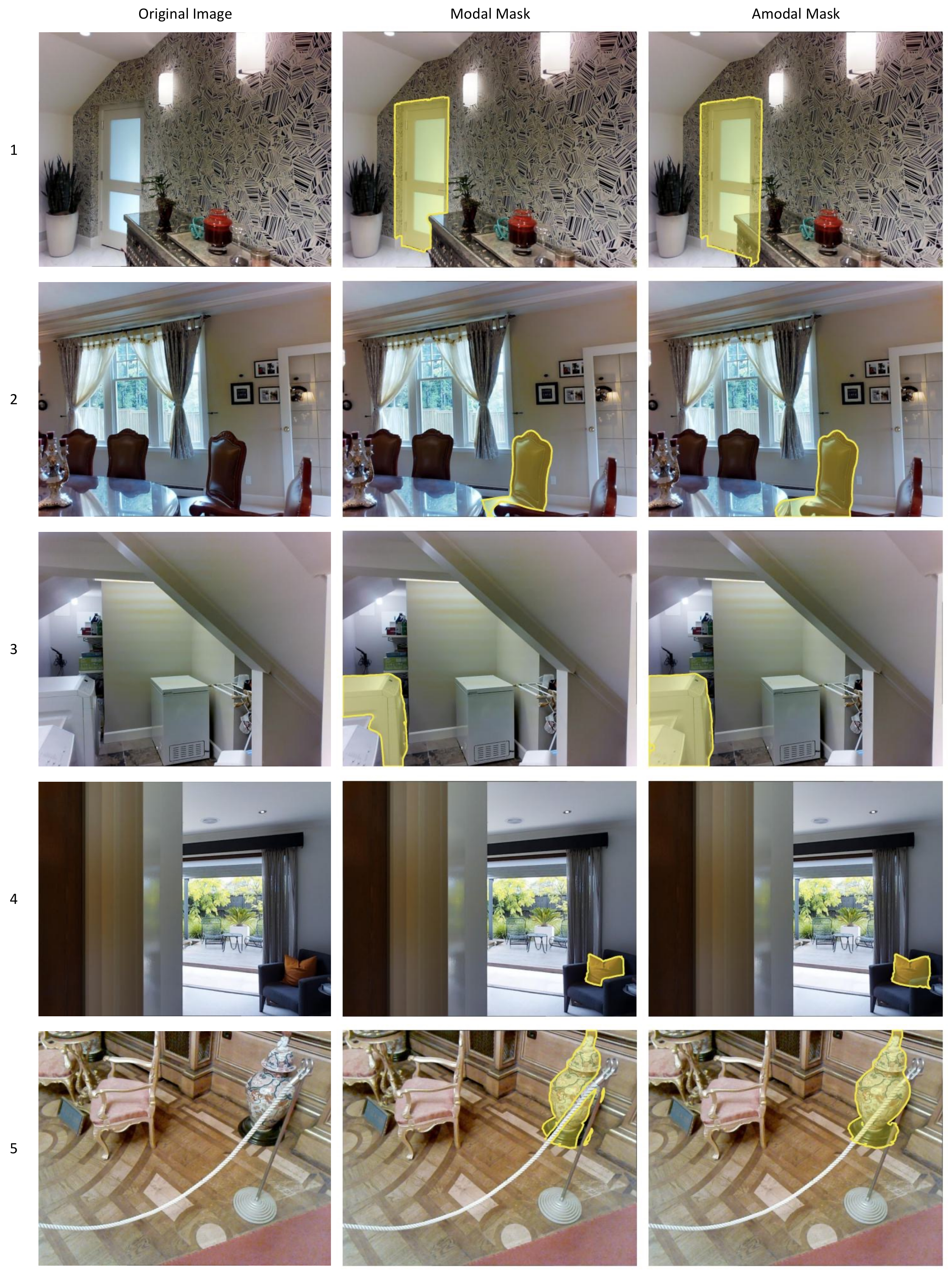}
		\caption{\textbf{More examples of our MP3D-Amodal Dataset}. For each example, original image together with generated modal and amodal masks are displayed.}
		\label{fig:sup_mp3d_example}
\end{figure*}

Figure~\ref{fig:sup_mp3d_example} shows more examples of our MP3D-Amodal dataset. For each example, original image together with generated modal and amodal masks are displayed.

\clearpage
\section{Manual Annotation of the MP3D-Amodal Dataset}
\label{sec:sup_manual_correction}

As described in Section~\ref{sec:generate_mp3d_dataset} of the main paper, modal and amodal masks for occluded objects are automatically selected via projecting all objects and only one object to the camera, and comparing the obtained modal and amodal mask. But not all generated modal and amodal masks are of very good quality, because the MatterPort3D meshes can be noisy and incomplete sometimes. Therefore, we ask human annotators to select the masks of good quality from all proposals.

We use the VGG Image Annotator (VIA)~\cite{dutta2019vgg} (Section~\ref{sec:sup_via_tools}) for the manual selection process of collecting our \emph{MP3D-Amodal}. The overview of the manual annotation steps and our user guide are in Section~\ref{sec:sup_user_guide}. 

\subsection{VIA Tools}
\label{sec:sup_via_tools}

VGG Image Annotator (VIA)~\cite{dutta2019vgg} is an open source manual annotation tool developed at the Visual Geometry Group. It can simply run in a web browser to collect answers from annotators for image, and therefore we use it for the manual annotation of our collected \emph{MP3D-Amodal} dataset.

\subsection{User Guide}
\label{sec:sup_user_guide}

\begin{figure*}[!htb]
		\centering
		\includegraphics[height=1.25\linewidth]{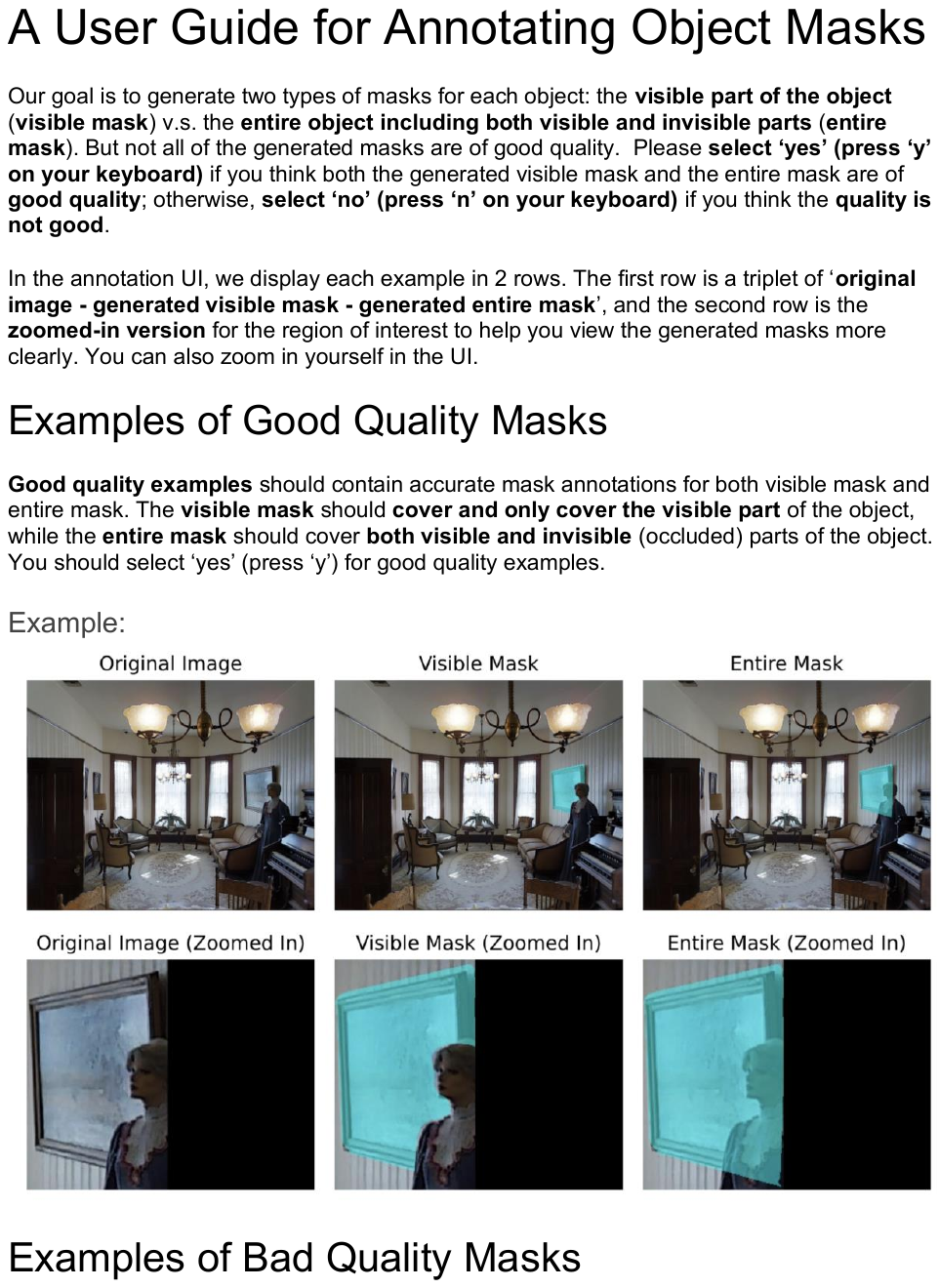}
		\vspace{-2mm}
		\caption{\textbf{User Guide for MP3D-Amodal Dataset Manual Selection (Page 1). 
  }}
		\label{fig:sup_via_user_guide_1}
 		\vspace{-4mm}
\end{figure*}

\begin{figure*}[!htb]
		\centering
		\includegraphics[trim=0.5cm 0cm 0.5cm 0cm, height=1.25\linewidth]{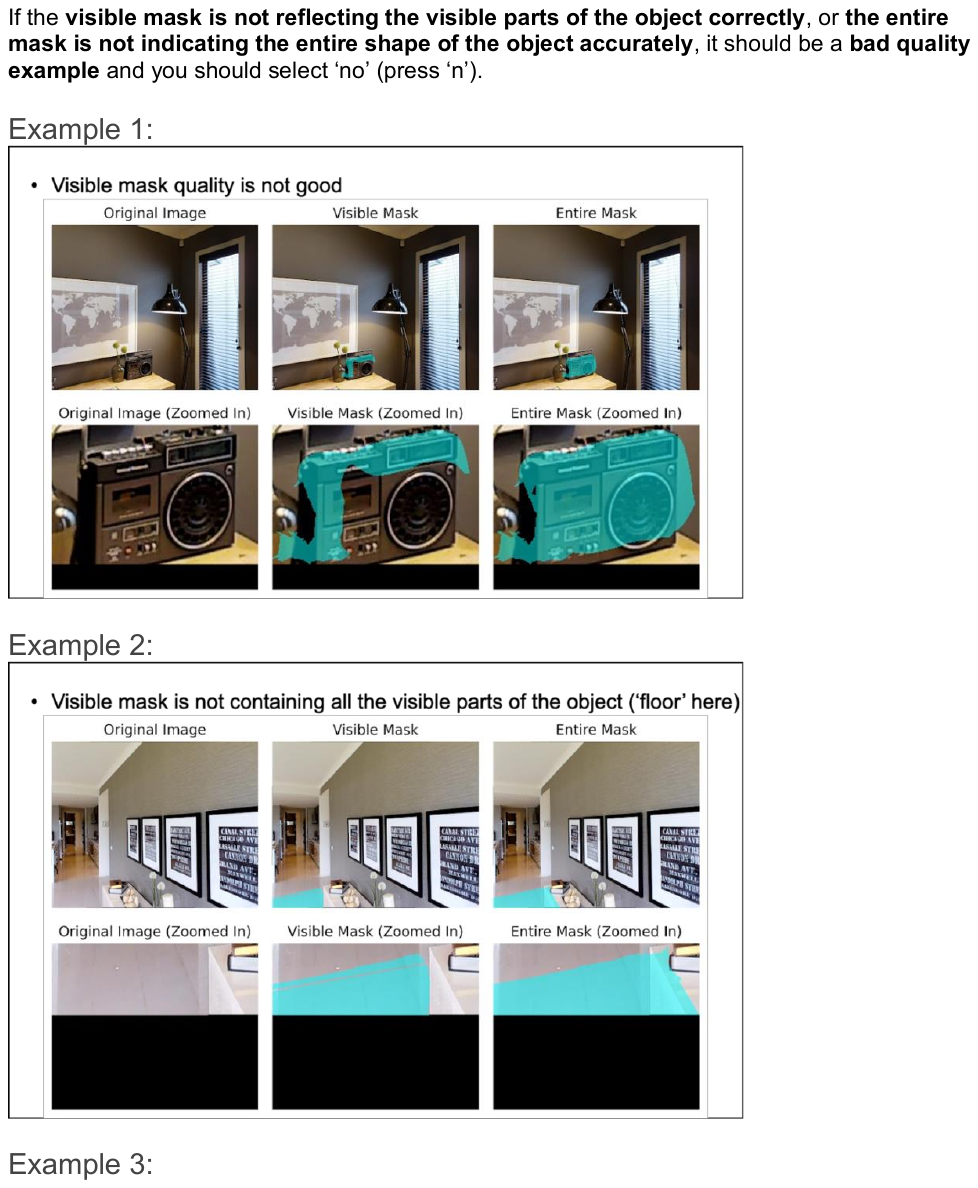}
		\vspace{-2mm}
		\caption{\textbf{User Guide for MP3D-Amodal Dataset Manual Selection (Page 2).} 
  }
		\label{fig:sup_via_user_guide_2}
 		\vspace{-4mm}
\end{figure*}

\begin{figure*}[!htb]
		\centering
		\includegraphics[trim=0.5cm 0cm 0.5cm 0cm, height=1.25\linewidth]{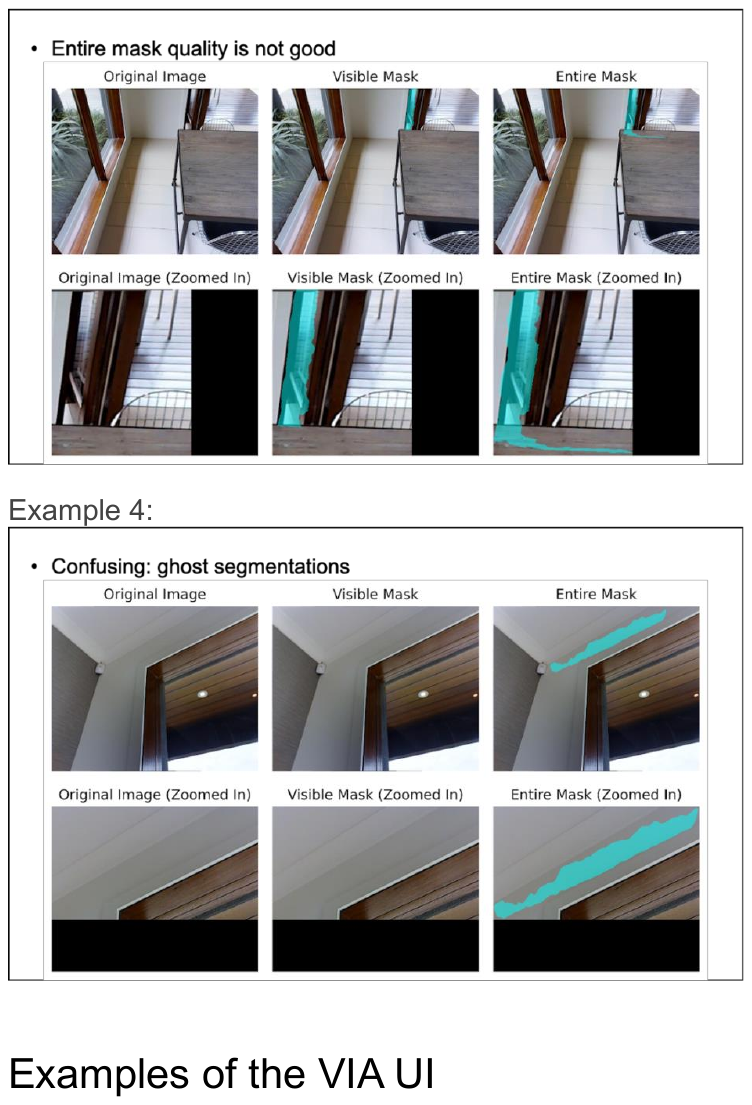}
		\vspace{-2mm}
		\caption{\textbf{User Guide for MP3D-Amodal Dataset Manual Selection (Page 3).}}
		\label{fig:sup_via_user_guide_3}
 		\vspace{-4mm}
\end{figure*}

\begin{figure*}[!htb]
		\centering
		\includegraphics[trim=0.5cm 0cm 0.5cm 0cm, height=1.25\linewidth]{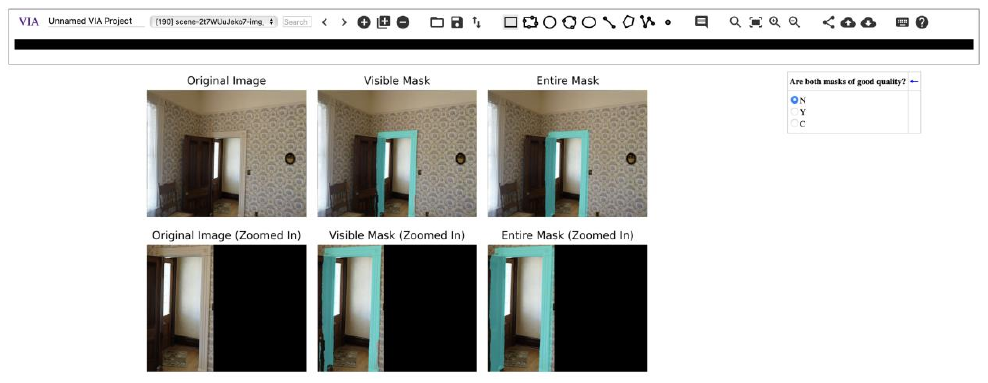}
		\vspace{-2mm}
		\caption{\textbf{User Guide for MP3D-Amodal Dataset Manual Selection (Page 4).}}
		\label{fig:sup_via_user_guide_4}
 		\vspace{-4mm}
\end{figure*}

For each generated modal and amodal masks, we display them together with the original image to the annotators, and ask them to answer 'Yes' or 'No' for the question "Are both masks of good quality?". See Figures~\ref{fig:sup_via_user_guide_1},~\ref{fig:sup_via_user_guide_2},~\ref{fig:sup_via_user_guide_3},~\ref{fig:sup_via_user_guide_4} for our user guide as the full text of instructions given to the annotators, including the screenshot of our VIA web page. 
There are two rounds of manual selection, where in the second round expert annotators further filter the `Yes' examples in the first round, to ensure that the selected masks are of good quality.
Annotators are paid more than the minimum salary in our country.

\end{document}